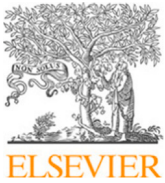

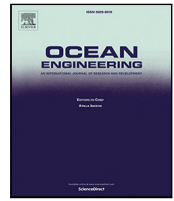

# Ship performance monitoring using machine-learning

Prateek Gupta [a,*], Adil Rasheed [b], Sverre Steen [a]

[a] Norwegian University of Science and Technology (NTNU), Department of Marine Technology, Trondheim, 7052, Sør-trondelag, Norway
[b] Norwegian University of Science and Technology (NTNU), Department of Engineering Cybernetics, Trondheim, 7034, Sør-trondelag, Norway

## ARTICLE INFO

## ABSTRACT

The hydrodynamic performance of a sea-going ship varies over its lifespan due to factors like marine fouling and the condition of the anti-fouling paint system. In order to accurately estimate the power demand and fuel consumption for a planned voyage, it is important to assess the hydrodynamic performance of the ship. The current work uses machine-learning (ML) methods to estimate the hydrodynamic performance of a ship using the onboard recorded in-service data. Three ML methods, NL-PCR, NL-PLSR and probabilistic ANN, are calibrated using the data from two sister ships. The calibrated models are used to extract the varying trend in ship's hydrodynamic performance over time and predict the change in performance through several propeller and hull cleaning events. The predicted change in performance is compared with the corresponding values estimated using the fouling friction coefficient ($\Delta C_F$). The ML methods are found to be performing well while modeling the hydrodynamic state of the ships with probabilistic ANN model performing the best, but the results from NL-PCR and NL-PLSR are not far behind, indicating that it may be possible to use simple methods to solve such problems with the help of domain knowledge.

## 1. Introduction

The hydrodynamic performance of a ship is an important factor which must be considered while working towards the green shipping future. In the ongoing development towards low-emission shipping, sparked partly by the IMO's goal of 50% reduction of greenhouse gas (GHG) emissions from global shipping within 2050, energy saving is going to be much more important, also from an economic perspective. Optimizing the hydrodynamic performance of a ship would not only lead to direct reduction of GHG, when running the ship on fossil fuels, but would also make alternative low-GHG fuel options more economically viable, as all the low-GHG alternative energy sources are significantly more expensive per unit energy than the traditional marine fuels.

The operational efficiency and, therefore, fuel or energy savings can be increased by keeping the hull and propeller smooth and clean of marine fouling (Townsin, 2003). However, cleaning very frequently is quite expensive and may also lead to increase in wear of the hull coating, which in turn may increase the resistance and fuel consumption (Munk, 2016). Conventionally, most of the ship owners perform scheduled maintenance (hull and propeller cleaning) on a regular timely basis which may not be very efficient. If the performance of the in-service ship can be efficiently and accurately monitored, the hull and propeller maintenance intervals can be optimized.

A sea-going ship is, now-a-days, equipped with numerous sensors which are continuously recording several variables, some of them representing the hydrodynamic state of the ship. These in-service recorded variables can be used to monitor the hydrodynamic performance of the ship (as suggested by Pedersen and Larsen, 2009). The objective of the current work is to use machine-learning (ML) methods to monitor the hydrodynamic performance of a ship over time using the in-service data recorded onboard the ship. Pedersen and Larsen (2009) used an artificial neural network (ANN) with just one hidden layer to model the hydrodynamic state of a ship, indicating the simplicity of the problem. Therefore, it may be possible to solve such a problem using a simple interpretable model like Principal Component Regression (PCR) or Partial Least Squares Regression (PLSR) with the help of some non-linear transformations obtained from our domain knowledge, thereby, linearizing the problem. In case of a linear problem, a simple model like PLSR is known to have outperformed ANN (Farifteh et al., 2007).

The current work focuses on developing data-driven methods for ship performance monitoring using the in-service data recorded onboard two sister ships. It is attempted here to predict the performance of the ships using two well-known multivariate linear regression models, namely, Principal Component Regression (PCR) and Partial Least Squares Regression (PLSR), enhanced using simple (but approximate) non-linear transformations, obtained from our domain knowledge. The enhanced models are used alongside an advanced probabilistic non-linear model, probabilistic artificial neural network (ANN), so that the performance of the enhanced models can be compared with the state-of-the-art. Moreover, a fouling growth factor is proposed to account for the growth of fouling on the ships' hull and propeller. The calibrated models are used to observe the fitted trend in the hydrodynamic state variables and predict the change in performance of the ships over time through several propeller and hull cleaning events. The predicted change is performance is compared with the corresponding values obtained from the fouling friction coefficient ($\Delta C_F$). Finally, the evolution of calm-water speed-power curve over time for the given ships is predicted.

---

* Corresponding author.
*E-mail addresses:* prateek.gupta@ntnu.no (P. Gupta), adil.rasheed@ntnu.no (A. Rasheed), sverre.steen@ntnu.no (S. Steen).

The following section describes the ML methods used here to model the hydrodynamic state of a ship. Section 3 contains a detailed literature survey regarding marine fouling and the formulation of the fouling growth factor used to include the effect of fouling growth on the hull and propeller of the ships. The datasets used to calibrate and validate the ML models are presented in Section 4. Lastly, Sections 5–7 presents the final results, conclusion and possible future work, respectively.

## 2. Machine-learning (ML)

Machine-learning (ML) is a broad subject involving design and operations of a diverse set of algorithms based on statistical methods. The main purpose of a statistical model is to draw inference, i.e., trying to establish the relationship between different variables, thereby helping physics establish or verify empirical relationships. ML is based on the same principle, but it serves a different purpose. ML models are designed with focus on predictive capabilities. In order to improve the predictive capabilities of ML models, simple and interpretable statistical models are transformed into highly complex and esoteric algorithms, like neural networks. With this adaptation, some ML models becomes more obscure (leading to the creation of so-called 'black-box' models), but it should be noted that ML also contains a substantial number of transparent and interpretable algorithms. These transparent algorithms can be used to solve a wide range of problems, but the simplicity of these transparent algorithms sometimes cannot compete with the complex algorithms due the complex nature of the problem-at-hand. Thus, it should be the role of the user to find a balance between interpretability and performance while doing method selection for a complex problem. In view of above, the current work presents three different ML models with varying complexity, interpretability and performance.

### 2.1. Regression modeling: Method & variable selection

Regression modeling is a statistics-based technique mainly used for inference and prediction. In other words, it is used to estimate the relationship between the target and input variable(s). It can thereafter be used to predict the target variable(s) from the input variable(s). One of the simplest cases of regression modeling is linear regression between a regressor (or input variable) and a regressand (or target variable). In such a simple case, it is possible to just plot the regressand against the regressor to ensure that there exists a good linear correlation between them. The performance of the regression model depends on how strong is this correlation.

In some cases, it is possible that the variation in the regressand(s) cannot be completely explained by the variation in the regressor(s), i.e., the given regressand(s) may also be dependent on some other unidentified regressor(s). Thus, the predictive performance of any regression model at least depends on the following two factors: (a) Correlation between the regressands and regressors; (b) The amount of variance in the regressands explained by the regressors. Further, it is naturally understandable that including an uncorrelated variable as an additional regressor cannot increase the model performance, rather in some cases, it may unnecessarily increase the model size and, therefore, the required computational time. Thus, variable selection is an important step in regression modeling. This is here achieved by Principal Component Analysis (PCA) (as demonstrated by Gupta et al., 2019).

PCA is a very powerful statistical method which can be used to fulfill several different purposes at the same time. In other words, PCA can be used as a one-stop shop for basic statistical analysis. One of the biggest advantage of PCA is that it is a completely transparent and interpretable ML method. Moreover, it can be further extended to a multivariate linear regression model, known as Principal Component Regression (PCR) (Jolliffe, 2002). PCR is different from ordinary least squares (OLS) linear regression as it, first, factorizes the regressor (or input) matrix into Principal Components (PCs), and then, uses only the first few PCs to carry-out the least squares regression. Thus, the PCR may produce better results as it reduces the noise in the regressors, by way of filtering-out the last few insignificant PCs, and the fact that the PCs are orthogonal to each other eliminates the problem of multicollinearity (well-known in the case of OLS regression). It should also be noted that PCR with maximum possible PCs (equal to the rank of regressor matrix) is exactly same as the OLS linear regression.

As mentioned above, PCR is a linear regression method, but it is well-known from our domain knowledge that the problem of ship propulsion is non-linear in nature. Moreover, one of the short comings of PCR is that the PC factorization of the regressor matrix is done independently from the regressands. Therefore, the sequence of PCs produced by the factorization of the regressor matrix is not always in the desired order of diminishing correlation with the regressands (Martens and Martens, 2001). This is generally resolved using a slightly modified method known as Partial Least Squares Regression (PLSR). On the other hand, to handle the non-linearities, some non-linear transformations can be used, thereby converting a non-linear problem into a linear one, and transforming PCR and PLSR into non-linear PCR (NL-PCR) and non-linear PLSR (NL-PLSR), respectively. In order to achieve the best possible results using such a method the user must be able to model all the non-linear dependencies. This is sometimes not feasible due to the complexity of the problem. Alternatively, it is possible to use a more complex non-linear model like artificial neural network (ANN) in order to better model the non-linear nature of the problem.

Considering the above, the following three supervised ML algorithms are used to create statistics-based regression models for the current work: (a) Principal Component Regression (PCR); (b) Partial Least Squares Regression (PLSR); (c) Probabilistic artificial neural network (Probabilistic ANN).

### 2.2. Principal component regression (PCR)

PCR is a linear regression model, based-on Principal Component Analysis (PCA), in which the regressors (or input variables) are the first few significant Principal Components (PCs). The mean-centered and standardized input data matrix ($X$) is, first, factorized using PCA as per the following equation:

$$\mathbf{X}^{m \times n} = \mathbf{T}_A^{m \times A} \cdot \mathbf{P'}_A^{A \times n} + \mathbf{E}_A^{m \times n} \tag{1}$$

Where $\mathbf{T}_A$ is PC scores' matrix and $\mathbf{P}_A$ is PC loadings' matrix with each column corresponding to a PC, and $\mathbf{E}_A$ is the residual matrix. Superscripts represent the dimensions of the matrices, i.e., $m$ is the number of samples, and $n$ is the number of input variables. $\mathbf{P'}_A$ represents the transpose of $\mathbf{P}_A$. $A$ is the model dimensionality or number of PCs. Further, the target variables are regressed using a linear regression model on the selected set of PC scores ($T_A$) as follows:

$$\mathbf{Y}^{m \times k} = \mathbf{T}_A^{m \times A} \cdot \mathbf{B}^{A \times k} + \mathbf{e}^{m \times k} \tag{2}$$

Where $\mathbf{Y}$ is the target matrix with $k$ target variables, $\mathbf{B}$ is the regression coefficient matrix and $\mathbf{e}$ is the regression residual matrix. Ridge regression with built-in cross-validation (*RidgeCV* from *Scikit-learn* in python (Pedregosa et al., 2011)) is used here for linear regression. Finally, to account for the non-linearities, additional variables are appended to the input ($X$) and target ($Y$) matrices after applying simple non-linear transformations on one original variable at a time. The PCR model enhanced with these non-linear transformations is further referred to as NL-PCR.

## 2.3. Partial least squares regression (PLSR)

PLSR is also a linear regression model based on the similar principle as PCR. Like PCR, it involves 2 steps: (1) Establishing a score matrix ($T_A$); (2) Regressing Y on $T_A$. But, in PLSR, $T_A$ is derived from both the regressor ($X$) and regressand ($Y$), and it is done in a manner such that $t_a$ vectors (comprising $T_A$) are produced in the desired order of diminishing correlation with the regressands ($Y$). Here, a simple iterative algorithm based on NIPALS (Vandeginste et al., 1988) is carried-out as follows:

$$w_a = \frac{E'_{a-1} \cdot u_a}{\|E'_{a-1} \cdot u_a\|} \quad \text{and} \quad t_a = E_{a-1} \cdot w_a \tag{3}$$

$$q_a = \frac{u'_a \cdot t_a}{\|u'_a \cdot t_a\|} \quad \text{and} \quad u_a = F_{a-1} \cdot q_a \tag{4}$$

Where $u_a$ is chosen as any one column of $F_{a-1}$, and $E_{a-1}$ and $F_{a-1}$ are $X$ and $Y$ residuals, respectively, obtained after extracting $(a - 1)$ factors, therefore, $E_0 = X$ and $F_0 = Y$. The above steps are carried-out iteratively until $t_a$ converges. The $X$ loadings ($p_a$) are calculated by using the relation indicated in Eq. (1) as follows:

$$p_a = \frac{E'_{a-1} \cdot t_a}{\|t'_a \cdot t_a\|} \tag{5}$$

Further, $X$ and $Y$ residuals are calculated as follows, and the process is repeated to extract the next set of scores and loadings.

$$E_a = E_{a-1} - t_a \cdot p'_a \quad \text{and} \quad F_a = F_{a-1} - u_a \cdot q'_a \tag{6}$$

For the current work, the PLSR model is created using the *PLSRegression* function defined by *Scikit-learn* in python (Pedregosa et al., 2011), and the non-linear variables are introduced in the same manner as explained above for the PCR model. The PLSR model enhanced with the non-linear transformations is further referred to as NL-PLSR.

## 2.4. Artificial neural network (ANN)

Originally inspired by the human brain, neural networks have proven to be excellent estimators for both classification and regression problems. Neural networks have out-competed many traditional signal processing and pattern recognition methods, becoming state-of-the-art in research fields, such as natural language processing (NLP) and computer vision (Schmidhuber, 2015). The simplest form of a neural network dates back to 1960s, called a multilayer perceptron (MLP) or a deep feed forward network (Schmidhuber, 2015). A MLP takes an input ($X$) and maps it to an output ($Y$) as $Y = f(X; \theta)$, where $\theta$ denotes the network parameters. The mapping can be seen as $L$ nonlinear mappings applied in succession: $f(X) = f^L(f^{L-1}(...f^2(f^1(X))))$. Each non-linear mapping is referred to as a layer, and the number of layers is known as the depth of the network.

$$\mathbf{l}^i = \sigma^i(\mathbf{W}^i \cdot \mathbf{l}^{i-1} + \mathbf{b}^i) \tag{7}$$

The output of a layer, $\mathbf{l}^i = f^i(f^{i-1}(...f^2(f^1(X))))$, is computed as shown in Eq. (7). First, the output of the previous layer is multiplied with a weight matrix ($\mathbf{W}^i$), and then, a bias vector ($\mathbf{b}^i$) is added to the product. The resulting vector is then fed to an activation function ($\sigma^i$) which is typically of the type sigmoid, ReLU or tanh (Goodfellow et al., 2016). The neural network can be visualized as a graph where each node, called a neuron, outputs an element of a layer vector ($\mathbf{l}^i$). The first column of neurons simply output the value of the input vector

($X$). The edges represent multiplication of the output of a neuron with an element of the weight matrix of the next layer. The products going into a neuron are summed, a bias is added and the activation function is applied. Analogous to multivariate linear regression, the model is trained to obtain an estimate of the set of model parameters (weights and biases), $\theta = [\mathbf{W}^1 \quad \mathbf{W}^2 \quad \cdots \quad \mathbf{W}^L \quad \mathbf{b}^1 \quad \cdots \quad \mathbf{b}^L]^T$, which minimizes the cost function, considering the hypothesis $\mathbf{Y} = f(\mathbf{X}; \theta)$. The most commonly used cost function is Mean Squared Error (MSE) with $L_2$ regularization.

$$\mathbf{J}(\theta) = \frac{1}{m}(\hat{\mathbf{Y}} - \mathbf{Y})^T(\hat{\mathbf{Y}} - \mathbf{Y}) + \lambda \sum_{i=1}^{L}(\|\mathbf{W}^i\|^2 + \|\mathbf{b}^i\|^2) \tag{8}$$

Where $m$ is the number of samples in $\mathbf{Y}$, $\hat{\mathbf{Y}}$ is the model prediction for $\mathbf{Y}$, and $\lambda$ is the weight decay or regularization parameter.

### 2.4.1. Probabilistic ANN: Dropout as a Bayesian approximation

It is well-known that a deep neural network is susceptible to overfitting. Applying additional dropout layers while training the model helps avoid this problem (Srivastava et al., 2014). Dropout refers to dropping-out units in a neural network. The units to be dropped-out are chosen randomly with a probability, known as the dropout probability ($p_{drop}$). This is mathematically equivalent to multiplying a vector, obtained from a Bernoulli distribution with success rate $(1 - p_{drop})$, with the selected weight matrix ($\mathbf{W}^i$). Thus, the dropped-out units are not considered during a particular forward and backward pass. Once the model is trained, the dropout layers are removed, and the model can be used to obtain point predictions, hereafter referred to as *Standard dropout* predictions.

The reliability of predictions from a deep neural network model is also a well debated topic. Two main reasons for this debate are: (1) The esoteric or black-box nature of the model; (2) The lack of information about the uncertainty in the predictions. A statistics-based Bayesian model uses probability to represent all uncertainty within the model. In other words, a Bayesian model also estimates the uncertainty (or confidence) with which the model is making a prediction, thus, providing a better way for the end user to gauge the reliability of results. Gal and Ghahramani (2016) claimed that a neural network with a dropout layer applied before each network layer, named as dropout neural network, can be treated as a Bayesian approximation for the probabilistic deep Gaussian process (Damianou and Lawrence, 2012).

The dropout neural network used for the current work is based on the framework developed by Gal and Ghahramani (2016). The model uncertainty is obtained by performing $T$ stochastic forward passes through the already-trained network (with dropout layers still active) and averaging the results. Thus, calculating a Monte Carlo mean of the predictions (hereafter referred to as *MC dropout* predictions), along with an approximate predictive probability distribution.

$$\hat{\mathbf{Y}} = \frac{1}{T}\sum_{t=1}^{T}\hat{\mathbf{Y}}_t(\mathbf{X}, \mathbf{W}_t^1, \mathbf{W}_t^2, ...., \mathbf{W}_t^L) \tag{9}$$

The approximate predictive distribution can be further used to calculate the predictive variance. A small predictive variance corresponds to a strong or high confidence prediction, whereas a large variance represents high uncertainty in the model prediction.

### 2.5. Model selection

In case of NL-PCR and NL-PLSR models, it is crucial to decide the model dimensionality, i.e., the number of Principal Components (PCs, for NL-PCR) or factors (for NL-PLSR) included in the regression model. In this case, the model selection is done by "sequential mode" cross-validation (CV) scheme, proposed by Wold et al. (2001). Here, first, the dataset is divided into 20 slices or continuous folds, then, 20 CV models are trained with leave-one-fold-out scheme. The prediction residuals are calculated for each CV model only using the corresponding left-out fold. These prediction residuals from all the CV models are, then, collected to calculate the prediction residual sum of squares (PRESS), which represents the predictive ability of the model.

**Table 1**
Hyper-parameters for the probabilistic ANN model.

| Sl. no. | Hyper-parameter | Value |
|---|---|---|
| 1 | Optimization algorithm | Stochastic gradient descent (Adam) |
| 2 | Loss function | Mean Squared Error (MSE) |
| 3 | Activation function | Rectified Linear Unit (ReLU) |
| 4 | Batch size | 5% of the training samples |
| 5 | No. of epochs | 2000 |
| 6 | No. of stochastic forward passes (T) | 10,000 |
| 7 | Model precision ($\tau$) | 1.0 |
| 8 | Length scale ($l$) | 10.0 |
| 9 | Dropout probability ($p_{\text{drop}}$) | 0.2 |
| 10 | No. of hidden layers | 1 |
| 11 | No. of neurons in hidden layer | 50 |

Parallelly, a fitted residual sum of squares (SS) is calculated for each model dimensionality using the full model (trained on the full dataset). Finally, the ratio $\text{PRESS}_a/\text{SS}_{a-1}$ is calculated of each model complexity ($a$) and a component or factor is considered significant if this ratio is smaller than around 0.9 for at least one of the target variables. Once a non-significant component or factor is observed, all the further PCs or factors are dropped, and the model dimensionality is set to $a - 1$. It should be noted that, in the ratio $\text{PRESS}_a/\text{SS}_{a-1}$, $\text{PRESS}_a$ is calculated for the model with '$a$' number of PCs or factors, but $\text{SS}_{a-1}$ is calculated for the model with '$a - 1$' number of PCs or factors.

Model selection in the case of artificial neural networks (ANN) is a tedious process which involves searching for an optimum model in a multi-dimensional hyper-parameter space. In other words, it is required to find a set of hyper-parameter values which produces the optimum model. Table 1 shows the list of hyper-parameters selected here for the probabilistic ANN model. The model precision ($\tau$), length scale ($l$) and dropout probability ($p_{\text{drop}}$) are further used to calculate the regularization parameter ($\lambda$) as follows (Gal and Ghahramani, 2016):

$$\lambda = \frac{l^2(1 - p_{\text{drop}})}{2n\tau} \tag{10}$$

where $n$ is the number of training samples. As indicated above, some of these hyper-parameters are obtained by searching for an optimum model. Therefore, the last 4 hyper-parameters (in Table 1), from 8 to 11, are obtained after testing several model architectures over a grid of length scale and dropout probability (further explained in Section 5.2). The remaining hyper-parameters are assumed to be constant.

The optimization algorithm, loss function and activation function are set to stochastic gradient descent (Adam), MSE (Mean Squared Error) and ReLU (Rectified Linear Unit), respectively, as generally recommended for ANN models for regression-based problems (Goodfellow et al., 2016). Adam is a highly efficient optimization algorithm (Ruder, 2016) used to adaptively vary the learning rate over the epochs to ensure fast and smooth convergence. The two most popular activation functions used for regression-based problems are hyperbolic tangent (tanh) and ReLU. ReLU is known to have several advantages over the other, like avoiding the problem of vanishing gradients (Glorot et al., 2011).

## 3. Marine fouling

As discussed by Yebra et al. (2004), the growth rate of biofouling on marine structures is dependent on many factors like water temperature, salinity, solar radiation, water depth, nutrients, etc. Additionally, the ability of these organisms to remain attached to the hull depends on the shear force acting on the hull (and, consequently, the ship speed) and the type of anti-fouling coating, as demonstrated by Fabbri et al. (2018) in an experimental study. Unfortunately, with the current state of technological advancements, it is not practical to have continuous visual monitoring of the hull surface to determine the extent of marine growth on the ship.[1] Therefore, the extent of fouling growth is here estimated using the in-service data recorded onboard the ship.

Based on the final objective, most of the prominent research in this field can be broadly categorized into following two types of methods: (i) Detection of marine fouling; (ii) Finding trend in ship's performance. The first type of methods includes the use of anomaly detection methods (Coraddu et al., 2019a; Logan, 2012). These methods are, unfortunately, only demonstrated to differentiate between the clean hull and a fouled one. They seem to be unable to quantify the extent of fouling growth. The second type of methods includes studies where researchers observed a trend in ship performance varying over time. In these methods, the authors used one or several of the following parameters, representing the performance of the ship: (a) Increased power demand (Walker and Atkins, 2007; Ejdfors, 2019); (b) Speed-loss (Koboević et al., 2019; Coraddu et al., 2019b); (c) Admiralty coefficient (Ejdfors, 2019); (d) Resistance or resistance coefficient (Munk, 2016; Foteinos et al., 2017; Ejdfors, 2019).

Ejdfors (2019) presented a comparison between three of the above mentioned alternatives (a, c and d) and concluded that using the admiralty coefficient produces the most logically-correct results.[2] Ejdfors (2019) assumed the admiralty coefficient to be constant for a given ship over a range of speed-power-displacement, therefore, it could be perceived as a summarized calm-water speed-power curve for a given displacement. Gupta et al. (2021) analyzed this assumption and concluded that this assumption only holds good for a generalized form of admiralty coefficient as the speed and displacement exponents in the original admiralty coefficient are not valid for modern hull forms. Gupta et al. (2021), therefore, proposed to use the generalized admiralty coefficient ($\Delta^m V^n/P_s$) as the statistical hydrodynamic performance indicator for a ship, with displacement and speed exponents, i.e., $m$ and $n$, respectively, obtained statistical using the in-service data recorded onboard the ship. The current work uses the generalized admiralty coefficient to formulate the variable used to account for the growth of marine fouling on the ships over time.

### 3.1. Statistical modeling of marine fouling: Fouling growth factor

In order to account for the change in hydrodynamic performance of a ship due to fouling growth in a data-driven model, it is required to statistically model the extent of fouling growth on the ship's hull and propeller over time. It may be possible to do it using the factors affecting the fouling growth (listed by Yebra et al. (2004)) but that would require observing and recording all these factors over the data recording duration to a good enough accuracy. Moreover, a well-established formulation may be required to calculate the extent of fouling growth at any point of time based on these factors. In the absence of any such information, it may be possible to fit a simpler expression which may be able to account for fouling growth on ship's hull and propeller over time.

Malone et al. (1981) presented a computer program for Hull Performance Assessment Model (HPAM) which can be used to compare and optimize hull surface management practices. The HPAM evaluates the performance of a ship over time after taking into account the operational variation in the ship's resistance due to environmental loads and hull roughness. Here, Malone et al. (1981) estimated the total hull roughness as a sum of 4 contributing components: steel plate

---

roughness, coating system roughness, corrosion roughness, and fouling roughness. The hull roughness due to each of these components is quantified in terms of a mean apparent amplitude (MAA), representing the average peak-to-trough vertical distance in the given area of hull surface.

Malone et al. (1981) estimated the hull fouling roughness based on the effective life of antifouling coating, ship's static time in port and port fouling severities. It was assumed that no fouling occurs while the ship is underway (over 3 knots) based on the test results from rotating drums immersed in salt water in several different port environments.[3] Therefore, the change in hull roughness due to marine fouling or hull fouling roughness was estimated as a function of cumulative static time the ship is subjected to a given in-port fouling severity. In the current approach, fouling roughness is represented by a fouling growth factor (FGF) calculated as the cumulative static time ($t_{static}$) during which the ship's speed is expected to be less than 3 knots multiplied by a fouling growth rate (FGR) as follows:

$$\text{FGF} = \sum_i t_{static,i}.\text{FGR}_i \tag{11}$$

The fouling growth rate (FGR) is estimated as the trend in generalized admiralty coefficient (as demonstrated by Gupta et al. (2021)). It should be noted that two variables are calculated here to account for the fouling growth on the hull and propeller assuming the same fouling growth rate (FGR), and each time the hull or the propeller is cleaned the corresponding FGF is reset to zero, as shown later in Fig. 4. The total fouling growth factor (FGF), used to calibrate the models, is calculated as the sum of hull and propeller FGFs.

## 4. Data

The in-service data used for the current work is obtained from 2 sister ships. One of the sister ships, further mentioned as the original ship, is the same ship presented by Gupta et al. (2019, 2021), but the dataset from the original ship, used here, is recorded over a slightly longer duration (3 years). The dataset from the sister ship, further referred to as the sister ship, is recorded over a duration of about 2.5 years. The data used to calibrate the machine-learning models is an assimilation of in-service measurement data recorded onboard a ship and weather hindcast data.

### 4.1. Ship data

The ships are ~200 m long general cargo ship with installed capacity of approximately 10 MW (MCR[4]) supplied with Marorka Online[5] web application. The original and sister ship datasets are recorded over a duration of 3 years and about 2.5 years, respectively. In case of both the ships, each sample is obtained by averaging over a period of 15 min. Table 2 presents the categorized list of all the ship data variables recorded onboard the ships.

### 4.2. Weather hindcast data

The weather hindcast data is obtained from European Centre for Medium-Range Weather Forecast (ECMWF) (Copernicus Climate Change Service (C3S), 2017), Hybrid Coordinate Ocean Model (HYCOM) (Chassignet et al., 2007) and Copernicus Marine Environment Monitoring Service (CMEMS) (Global Monitoring and Forecasting Center, 2018). The ECMWF data, used for the current analysis, is obtained from

ERA5 HRES (High Resolution) climate reanalysis dataset. The spatial resolution of ERA5 HRES is 0.25°, and the temporal resolution is 1 h. ECMWF dataset is used to obtain northward and eastward wind speeds 10 m above the sea surface, significant wave height, mean wave period and mean wave direction. HYCOM dataset, used here, has a spatial resolution of 1/12° with a sampling frequency of 1 measurement per day, and CMEMS dataset has the same spatial resolution but a sampling frequency of 1 measurement per hour. HYCOM and CMEMS datasets are used here to obtain northward and eastward sea water speed for the original ship and sister ship, respectively. The weather data variables are linearly interpolated in space and time to ship's location using the available navigation data.

### 4.3. Data exploration & pre-processing

Fig. 1 presents the speed-through-water (log speed) vs shaft power from the raw data recorded over a period of 3 years onboard the original ship and about 2.5 years onboard the sister ship. The design speed of both the ships is 15.5 knots. In comparison to a typical calm-water curve, the raw data subplots (in Fig. 1) shows a good variation in power for a fixed speed as expected due variation in loading conditions as well as environmental loads. But nevertheless, this does not explain the samples with quite high shaft power at almost zero speed-through-water. A closer analysis reveals that such samples are obtained due to non-zero accelerations, i.e., periods when the ship is accelerating or decelerating.

Although it is possible to use these samples, but only after including a variable representing the rpm increase (or decrease) by the ship master, it is decided to remove these samples for the current work. After removing all the samples associated with non-zero accelerations (further referred to as unsteady samples), the ship can be assumed to be in a quasi-steady state at each observed sample. A quasi-steady filter, explained by Gupta et al. (2021), is applied to the shaft rpm and GPS speed time series to filter out these unsteady samples.

Fig. 2 shows the main steps of data pre-processing and validation. Here, the draft and trim measurements are corrected for Venturi effect, as explained by Gupta et al. (2021), and the hindcast validation is performed as explained in the following section. The processed data is further used to calibrate, validate and test the machine-learning models.

### 4.4. Data validation

*Speed-through-water or speed-over-ground?* It is often a question about whether to use speed-though-water (log speed) or speed-over-ground (GPS speed) for an analysis. It is well-known that the GPS speed measurements are more reliable, as the GPS sensors are very accurate these days, but the log speed is more relevant for a hydrodynamic analysis, as it represents the actual speed of the ship through the surrounding water which in-turn is directly correlated with the effective power at the propeller. The difference between the GPS and log speed is caused due to the longitudinal water current speed. In the absence of any water current, the GPS and log speeds are the same. The given ships are equipped with Doppler sonar sensors for log speed measurements. The Doppler-based log speed measurements may not be accurate enough to carry-out a hydrodynamic analysis, as pointed out by Dalheim and Steen (2021). Thus, it may be useful to include GPS speed measurements in the analysis along with the longitudinal water current speed from the hindcast.

*Weather hindcast data.* While using an external data source like hindcast, it is important to validate the interpolated data from the external source with the onboard recorded in-service data. Several such validations are carried-out in the current case as some of these hindcast variables are also recorded directly or in-directly onboard the ship. All the ships are generally equipped with anemometers, therefore, it is possible to validate the wind speed and direction obtained from

---

[3] It should be noted that some fouling may still occurs in case of full hull form ships (like tankers) sailing at slow speed but greater than 3 knots as the local flow speed, past certain parts of the hull, can be less than 3 knots (Malone et al., 1981).

[4] Maximum Continuous Rating of the main engine.

[5] www.marorka.com.

**Table 2**

Categorized list of variables recorded onboard the ships. Only 'Navigation', 'Propulsion System' & 'Environment' variables are used for the current analysis. *Abbreviations:* IMO = International Maritime Organization; COG = Center of Gravity; Aux. = Auxiliary; DG = Diesel Generator (for auxiliary power systems); ME = Main Engine (for propulsion system); GPS = Global Positioning System. * marked variables are only recorded for the original ship, whereas ** marked variables are only recorded for the sister ship.

| Ship identity | Navigation | Auxiliary power system | Propulsion system | Environment |
|---|---|---|---|---|
| Ship Name | Latitude | Aux. Consumed | State | Relative Wind |
| IMO Number | Longitude | Aux. Electrical | ME Load | Speed |
| | Gyro Heading | Power Output | Measured | Relative Wind |
| | COG Heading | DG1 Power | Shaft Power | Direction |
| | | DG2 Power | Shaft rpm | Sea Depth* |
| | | DG3 Power | Shaft Torque | Under Keel |
| | | | ME Consumed | Clearance** |
| | | | Draft Fore | |
| | | | Draft Aft | |
| | | | GPS Speed | |
| | | | Log Speed | |
| | | | Cargo Weight | |

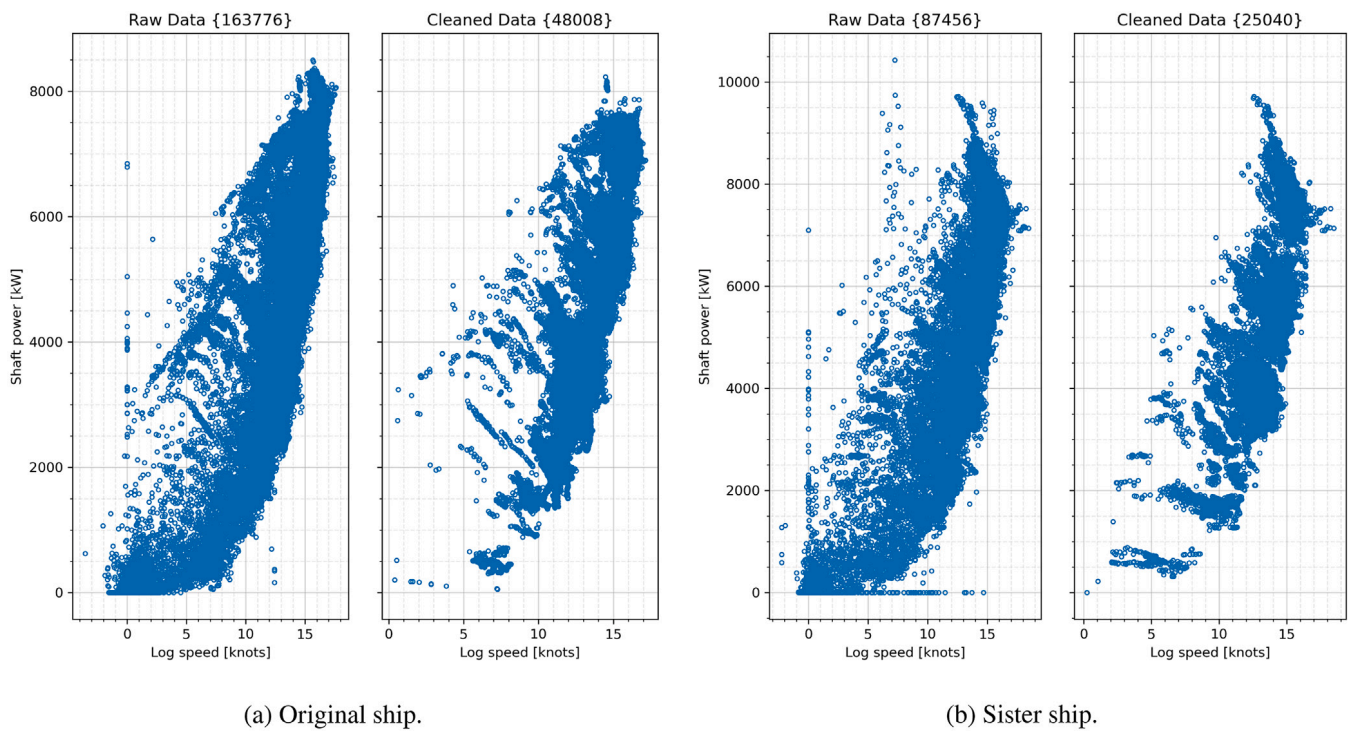

(a) Original ship.                    (b) Sister ship.

**Fig. 1.** Measured speed-through-water (or log speed) vs measured shaft power obtained from both the ships. The subplot on the left shows the raw data, and the subplot on the right shows the samples remaining after data cleaning. The numbers shown in curly brackets ({}) in the title of each subplot is the number of samples in the corresponding subplot.

the hindcast with the onboard measurements. Similarly, as mentioned above, the difference between the GPS and log speed (recorded onboard the ship) is the longitudinal water current speed, which is validated against the corresponding values obtained using the hindcast.

## 5. Results

This section presents the results obtained from the three regression models, NL-PCR, NL-PLSR and Probabilistic ANN. Table 3 shows the input and target variables used to calibrate these models. The data used to calibrate these models is obtained from two sister ship, as presented in Section 4. The datasets from both the ships is split into training (80%) and test (20%) datasets to check for model performance. In case of ANN, 10% of the training dataset is used for validation during model calibration. Further, the calibrated machine-learning models are used to predict the change in performance of both the ships over time through several propeller and hull cleaning events. The predicted change in performance is, finally, compared with the change in performance estimated using the fouling friction coefficient ($\Delta C_F$), calculated using the in-service data. The detailed procedure for calculating $\Delta C_F$ is presented in Appendix B.

### 5.1. Fouling growth factor (FGF)

As mentioned in Section 3.1, the fouling growth factor (FGF), included as an input variable (shown in Table 3) to represent the added resistance due to fouling growth on the hull and propeller of the ship, is calculated as the summation of two fouling growth factors, hull and propeller FGFs. The hull and propeller FGFs are calculated according to Eq. (11), where the fouling growth rate (FGR) is estimated using the trends in generalized admiralty coefficient, refer Gupta et al. (2021) regarding the detailed procedure for calculating the trends in the generalized admiralty coefficient. Appendix A presents the validation of the trends obtained using the generalized admiralty coefficient for both the ships by comparing them to the trends obtained from the traditional method, i.e., the fouling friction coefficient ($\Delta C_F$).

**Table 3**

Input and target variables used by the regression models, i.e., NL-PCR, NL-PLSR and Probabilistic ANN. * marked variables are only included in NL-PCR and NL-PLSR models. *Abbreviations:* Long. = Longitudinal; Trans. = Transverse; GPS = Global Positioning System.

| Sl. No. | Category | Variables |
|---|---|---|
| 1 | | Shaft rpm, Mean draft, Trim-by-aft |
| 2 | | Long. wind speed, Trans. wind speed, Long. current speed |
| 3 | Input | Significant wave height, Relative mean wave direction, Mean wave period |
| 4 | | Fouling growth factor |
| 5* | | Shaft rpm$^1$, Mean draft$^{1/2}$, Significant wave height$^2$ |
| 6 | Target | Shaft power, GPS speed, Log speed |
| 7* | | GPS speed$^3$, Log speed$^3$ |

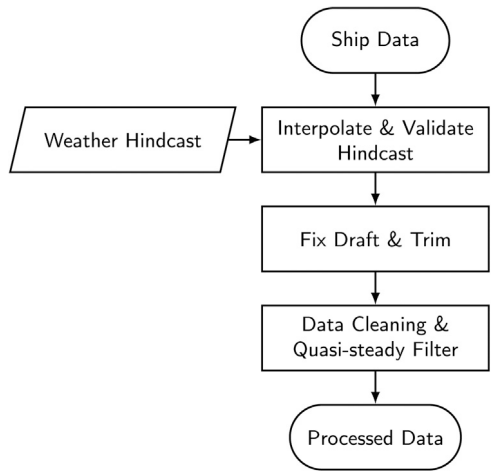

**Fig. 2.** Main data pre-processing and validation steps.

Ship Data → Interpolate & Validate Hindcast (← Weather Hindcast) → Fix Draft & Trim → Data Cleaning & Quasi-steady Filter → Processed Data

**Table 4**

Model dimensionality (number of PCs or factors) for PCR and PLSR models, obtained from the cross-validation results (shown in Fig. 5 for the sister ship).

| | NL-PCR | NL-PLSR |
|---|---|---|
| *Original Ship* | 5 | 4 |
| *Sister Ship* | 8 | 3 |

Following the method given by Gupta et al. (2021), first, the generalized admiralty coefficient ($\Delta^m V^n / P_s$) is obtained for the ships by statistically fitting the displacement and speed exponents (i.e., $m$ and $n$, respectively) to the near-calm-water filtered and corrected in-service data, and then, the trend lines are fitted between each propeller and/or hull cleaning event (as shown in Fig. 3). It should be noted that each data point in Fig. 3 represents the mean of all the generalized admiralty coefficient values calculated for an individual voyage (or part of a voyage), obtained at the same ship static time. Therefore, it should not be surprising that some of the trends observed here are opposite to the expectations, as the data is noisy and the methods used to correct the filtered data for near-calm-water conditions, i.e., added wind and wave resistance estimation methods, are known to have some uncertainty. Nevertheless, for calculation purpose, the abnormal trends observed in leg 1 and 6 for the original ship, and leg 2 and 6 for the sister ship (refer Fig. 3) were replaced by the least growth trend (observed in leg 7 for the original ship and leg 3 for the sister ship) so that the resulting FGFs remain logical.

Fig. 4 shows the hull and propeller FGFs calculated for the given time-series. The propeller of both the ships was cleaned several times during the data recording duration (marked by red vertical lines in Fig. 4), therefore, the propeller FGF is reset to zero just after each of these cleaning activities. Moreover, the hull was never cleaned for the original ship, but it was cleaned once for the sister ship (marked by the dashed black vertical line in Fig. 4(b)) during the data recording duration. Therefore, the hull FGF remains monotonically increasing for the original ship but is reset to zero once for the sister ship.

### 5.2. Regression modeling

*Model selection.* For NL-PCR and NL-PLSR models, the model selection is done using the "sequential mode" cross-validation (CV), as explained in Section 2.5. Fig. 5 shows the cross-validation results for the sister ship. From the results of CV, the model dimensionalities, i.e., the number of Principal Components (PCs) in NL-PCR or factors in NL-PLSR, are obtained as follows (also shown in Table 4): (a) Original ship: 5 PCs for NL-PCR and 4 factors for NL-PLSR; (b) Sister ship: 8 PCs for NL-PCR and 3 factors for NL-PLSR.

For ANN, the model selection is done keeping in mind the following factors: (a) Validation loss; (b) Difference between the training and validation loss; (c) Number of model parameters (i.e., the size of the network). In general practice, the model with minimum validation loss is chosen, as it indicates a good predictive performance, but the difference between the training and validation loss is also important as it indicates if the model is underfitted or overfitted. A good model should present comparable performance on both the training and validation datasets. Finally, the size of the network (or number of model parameters) represents the model complexity and computational resources required to calibrate the model.

Comparing 12 different network architectures with up to 5 hidden layers and 100 neurons (in a hidden layer) over a grid of dropout probability and model length scale, the model with just 1 hidden layer containing 50 neurons satisfied all the three above mentioned conditions as well as it performed very well on the datasets from both the ships. Thus, the final results here are presented based on a single hidden layer ANN model containing 50 neurons with 0.2 dropout probability and 10.0 length scale (as presented in Table 1).

*Training, validation and test split.* In all the cases, 80% of the data time-series is used for training the models, and 20% of the time-series is used to test the predictive performance of the models and, therefore, check for model overfitting and generalization. In case of ANN, as an additional check for model overfitting, 10% of the training set is used for validation while training the model. Although the 80/20 split, used here, is quite popular in machine-learning domain, there is no universal rule to decide the data splitting ratio. Some researchers use the 80/20 split based on the Pareto principle, which says that roughly 80% of consequences come from 20% of causes. Another way to decide the splitting ratio is using the scaling law introduced by Guyon (1997), but the scaling law does not seem appropriate to some researchers as it defines the splitting ratio based on the number of input (or independent) features instead of the number of samples and the quality of data. It seems more appropriate to split the data in such a manner that the samples in the training set contains all the important patterns which should be learned by the model, but it should not result in overfitting the model. Thus, as long as the training set contains enough samples to confidently estimate all the model parameters (weights and biases) and satisfy the above criterion, the exact ratio would not matter.

Furthermore, it is generally recommended to obtain the train-validation-test split using random picking, but in the case of propulsion data from cargo ships, it should be noted that this strategy would most probably not be satisfactory. Cargo ships are generally propelled at an unchanged rpm or speed, in slowly varying weather conditions, for substantial durations of time, thereby resulting in consecutive replicated data samples in the time-series. A random selection of validation and test sets would produce a vast majority of these replicated samples which are already present in the training set. The validation and test datasets are required to check for model overfitting and generalization. Therefore, it is required for validation and test sets to contain samples which are not replicated from the training set. Rather, the validation and test sets should contain a majority of new untouched samples which can check if the model has properly generalized over the whole range of all the input variables. Thus, the train-validation-test split is done here by slicing-out sections of the time-series.

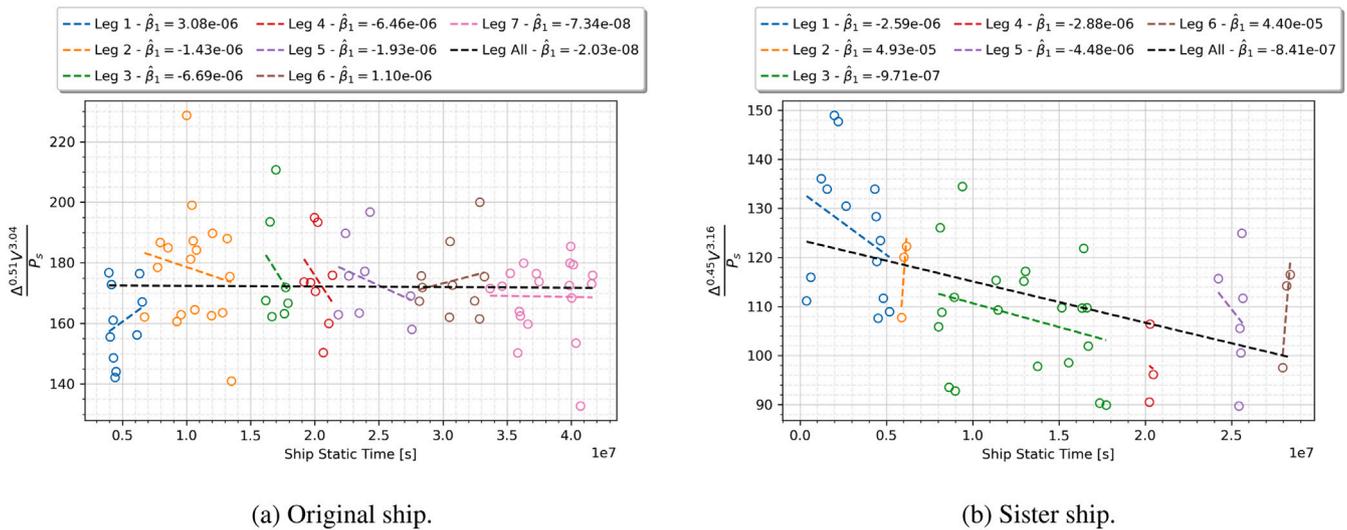

(a) Original ship.

(b) Sister ship.

**Fig. 3.** Trends in generalized admiralty coefficient used to estimate the fouling growth factors. (For interpretation of the references to color in this figure legend, the reader is referred to the web version of this article.)

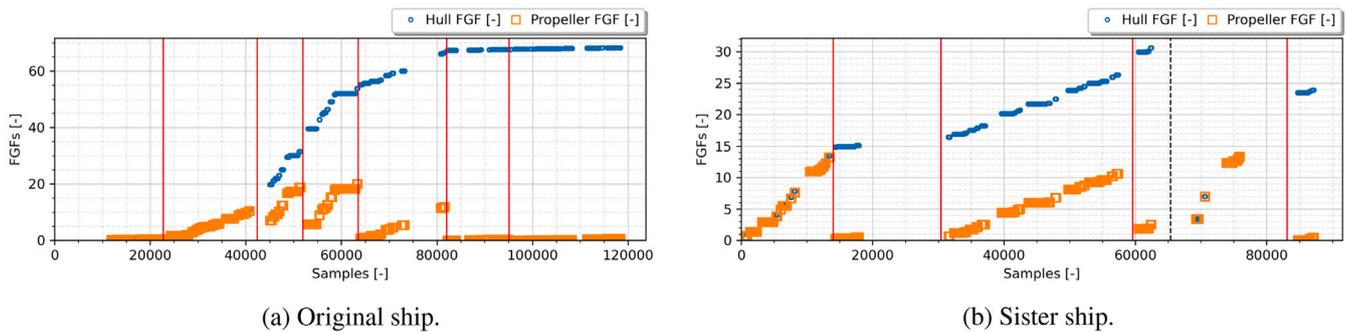

(a) Original ship.

(b) Sister ship.

**Fig. 4.** Fouling growth factors for hull and propeller calculated using Eq. (11). The red and dashed black vertical lines mark the propeller and hull cleaning events, respectively, occurring during the recorded time series. (For interpretation of the references to color in this figure legend, the reader is referred to the web version of this article.)

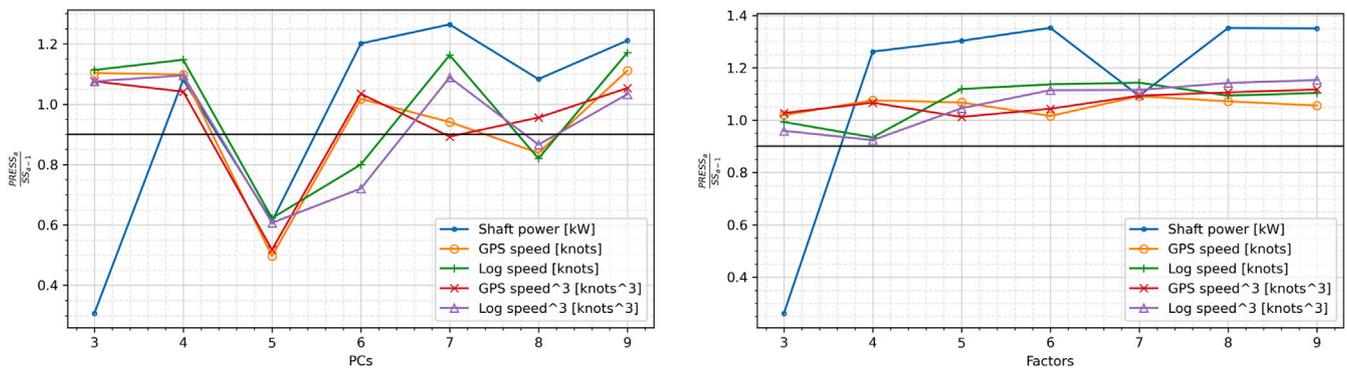

**Fig. 5.** NL-PCR and NL-PLSR model cross-validation results for the sister ship.

The current work aims to predict the change in performance of the ships through each propeller and hull cleaning event. Therefore, the training data should contain enough samples before and after each of these cleaning events in order to experience the change in performance through the corresponding event. Thus, the data splitting is done keeping in mind the occurrence of these cleaning events with respect to the time-series. Fig. 6 presents the data division for both the ship used to calibrate and test the regression models. For the original ship, one-third of the test data is taken from the starting of the time-series and the remaining two-third is taken from the end of the time-series. For the sister ship, one-fourth of the test data is taken from the starting and end of the time-series, and the remaining half of the test data is taken from the middle of the time-series. The validation data (only used in case of the ANN model) is taken from the middle of the training data time-series in both the cases.

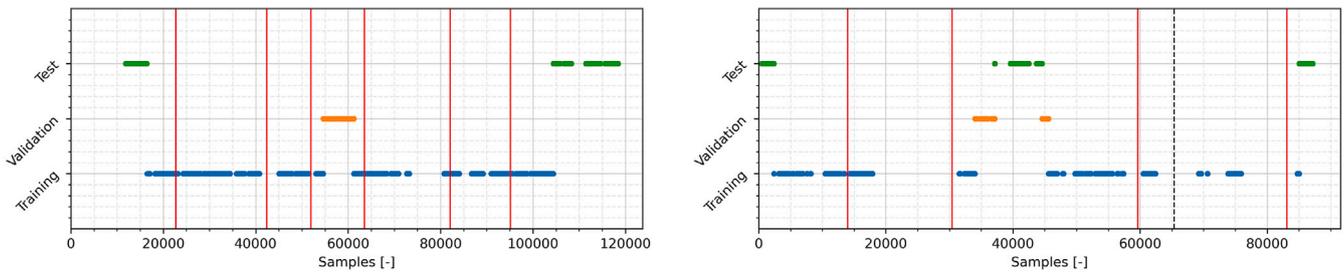

**Fig. 6.** Data division used to calibrate and test the regression models for the original ship (left) and sister ship (right). Validation is only done for the ANN model, therefore, the validation set remained as a part of the training set in the case of NL-PCR and NL-PLSR models. The red and dashed black vertical lines represents the propeller and hull cleaning events, respectively. (For interpretation of the references to colour in this figure legend, the reader is referred to the web version of this article.)

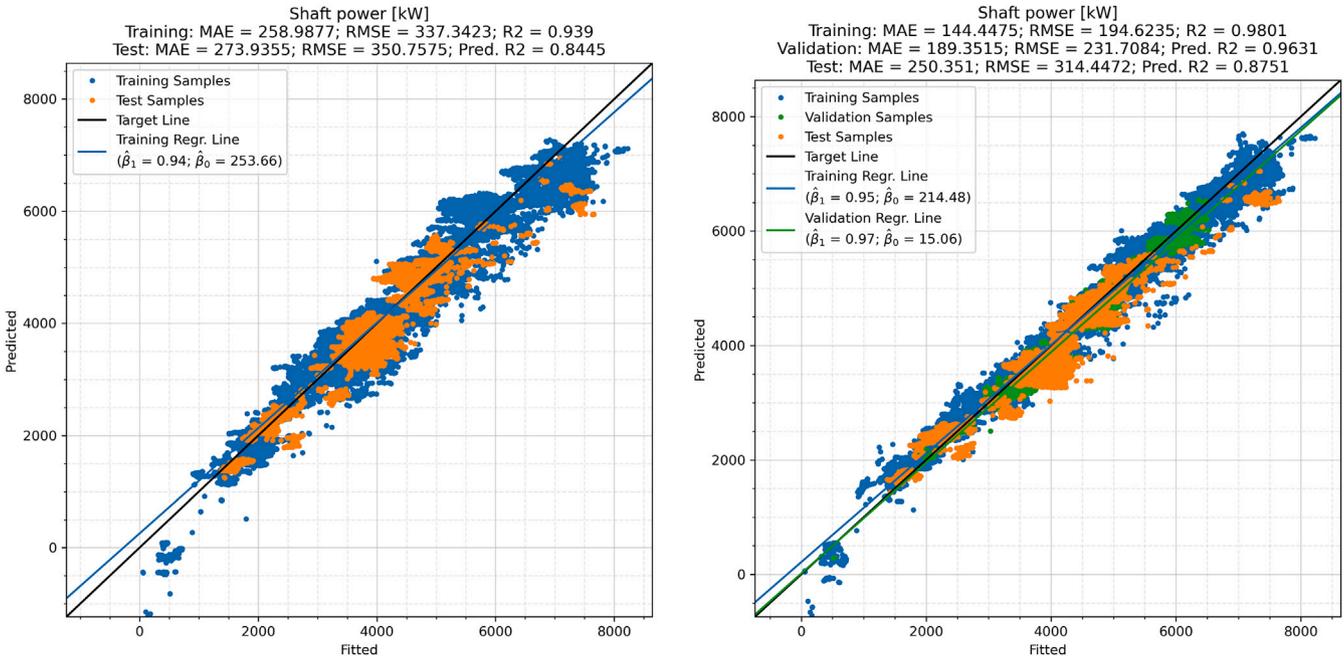

**Fig. 7.** Shaft power calibration results using NL-PLSR (left) and ANN (right) models for the original ship. (For interpretation of the references to color in this figure legend, the reader is referred to the web version of this article.)

*Model calibration and testing.* All the regression models are calibrated using the data obtained from both the ships, presented in Section 4. Table 3 presents the input and target variables used to calibrate the models. The goodness of fit is assessed using the following three well known parameters: MAE (Mean Absolute Error), RMSE (Root Mean Squared Error) and R2 (R-squared score or coefficient of determination). Table 5 shows the results for all three regression models. Fig. 7 shows the calibration results for the shaft power variable using the NL-PLSR and ANN models for the original ship. In all the cases, the ANN model presents the best performance, but the results from the NL-PCR and NL-PLSR models are not marginally different from the ANN model.

From Table 5, it seems like the non-linear transformed target variables (* marked), used in NL-PCR and NL-PLSR models, do not fit well, but a closer look at the prediction plots (refer Fig. 8) indicates that the poor RMSE and R2 score for these variables is due to a very poor fit for small numerical value samples. This is due to the fact that the data is sparse for small ship speeds (as evident from Fig. 1), and the non-linear transforms used in the NL-PCR and NL-PLSR models are, in fact, approximations, causing a not so well fit in the data tails. Nevertheless, the non-linear transformed target variables seems to be fitting well enough in the high speed range.

### 5.3. Ship's hydrodynamic performance trend

While doing machine-learning (ML), it is often considered a good practice to carry-out several generalization checks to understand the final state of the model. A ML model represents nothing but a mapping between the input and target variable space. Thus, it seems to be an interesting idea to question the calibrated model and understand the trends to which the model is fitted. The current section aims to use this idea in order to see the hydrodynamic performance trends learned by each of the currently used ML models.

Let us assume that the ships are being driven at a constant rpm, say, the NCR[6] rated rpm, in completely calm-water condition (no wind, wave and current loads) at a constant even draft (no trim) for the whole duration of the time-series. Now, it is first of all expected that the shaft power demand for the ships in the given condition would increase over time due to the growth of marine fouling. Further, it is expected that right after each cleaning event this power demand would suddenly drop. An opposite trend is expected from the ship speed. Here, both GPS and log speeds should give the same results as the water current is assumed to be absent. In order to check for these expected performance

---

**Table 5**

Calibration results for regression models for both the ships. The * marked target variables are the non-linear transformed variables (number 7 in Table 3) used in NL-PCR and NL-PLSR models. The 'ANN (MC dropout)' column presents the results using the Monte Carlo dropout estimates from the probabilistic ANN, as explained in Section 2.4.1.

| Target variable | NL-PCR | | | NL-PLSR | | | ANN | | | ANN (MC dropout) | | |
| --- | --- | --- | --- | --- | --- | --- | --- | --- | --- | --- | --- | --- |
| | MAE | RMSE | R2 | MAE | RMSE | R2 | MAE | RMSE | R2 | MAE | RMSE | R2 |
| **Original Ship - Training** | | | | | | | | | | | | |
| Shaft power [kW] | 245.81 | 311.06 | 0.95 | 258.99 | 337.34 | 0.94 | 144.45 | 194.62 | 0.98 | 144.46 | 194.64 | 0.98 |
| GPS speed [knots] | 0.61 | 0.83 | 0.78 | 0.48 | 0.66 | 0.86 | 0.39 | 0.53 | 0.91 | 0.39 | 0.53 | 0.91 |
| Log speed [knots] | 0.41 | 0.60 | 0.87 | 0.35 | 0.51 | 0.91 | 0.20 | 0.29 | 0.97 | 0.20 | 0.29 | 0.97 |
| GPS speed* [knots] | 0.63 | 1.08 | 0.62 | 0.54 | 1.22 | 0.52 | – | – | – | – | – | – |
| Log speed* [knots] | 0.45 | 1.08 | 0.59 | 0.39 | 1.15 | 0.54 | – | – | – | – | – | – |
| **Original Ship - Test** | | | | | | | | | | | | |
| Shaft power [kW] | 298.64 | 367.74 | 0.83 | 273.94 | 350.76 | 0.84 | 250.35 | 314.45 | 0.88 | 250.37 | 314.46 | 0.88 |
| GPS speed [knots] | 0.48 | 0.63 | 0.81 | 0.42 | 0.56 | 0.85 | 0.39 | 0.52 | 0.87 | 0.39 | 0.52 | 0.87 |
| Log speed [knots] | 0.37 | 0.48 | 0.87 | 0.28 | 0.38 | 0.92 | 0.23 | 0.30 | 0.95 | 0.23 | 0.30 | 0.95 |
| GPS speed* [knots] | 0.51 | 0.66 | 0.79 | 0.43 | 0.59 | 0.84 | – | – | – | – | – | – |
| Log speed* [knots] | 0.38 | 0.50 | 0.87 | 0.29 | 0.40 | 0.91 | – | – | – | – | – | – |
| **Sister Ship - Training** | | | | | | | | | | | | |
| Shaft power [kW] | 216.75 | 277.31 | 0.97 | 249.38 | 312.59 | 0.96 | 144.99 | 191.58 | 0.99 | 144.99 | 191.61 | 0.99 |
| GPS speed [knots] | 0.41 | 0.59 | 0.90 | 0.44 | 0.61 | 0.89 | 0.33 | 0.48 | 0.92 | 0.33 | 0.48 | 0.92 |
| Log speed [knots] | 0.35 | 0.53 | 0.91 | 0.41 | 0.59 | 0.89 | 0.23 | 0.36 | 0.95 | 0.23 | 0.36 | 0.95 |
| GPS speed* [knots] | 0.51 | 1.40 | 0.41 | 0.52 | 1.39 | 0.42 | – | – | – | – | – | – |
| Log speed* [knots] | 0.41 | 1.30 | 0.47 | 0.43 | 1.21 | 0.54 | – | – | – | – | – | – |
| **Sister Ship - Test** | | | | | | | | | | | | |
| Shaft power [kW] | 293.80 | 365.69 | 0.91 | 313.48 | 376.62 | 0.90 | 211.19 | 269.22 | 0.95 | 211.18 | 269.20 | 0.95 |
| GPS speed [knots] | 0.57 | 0.76 | 0.81 | 0.56 | 0.76 | 0.81 | 0.50 | 0.66 | 0.86 | 0.50 | 0.66 | 0.86 |
| Log speed [knots] | 0.52 | 0.72 | 0.81 | 0.57 | 0.85 | 0.73 | 0.39 | 0.59 | 0.87 | 0.39 | 0.59 | 0.87 |
| GPS speed* [knots] | 0.70 | 1.53 | 0.21 | 0.68 | 1.47 | 0.27 | – | – | – | – | – | – |
| Log speed* [knots] | 0.58 | 1.42 | 0.26 | 0.55 | 0.87 | 0.72 | – | – | – | – | – | – |

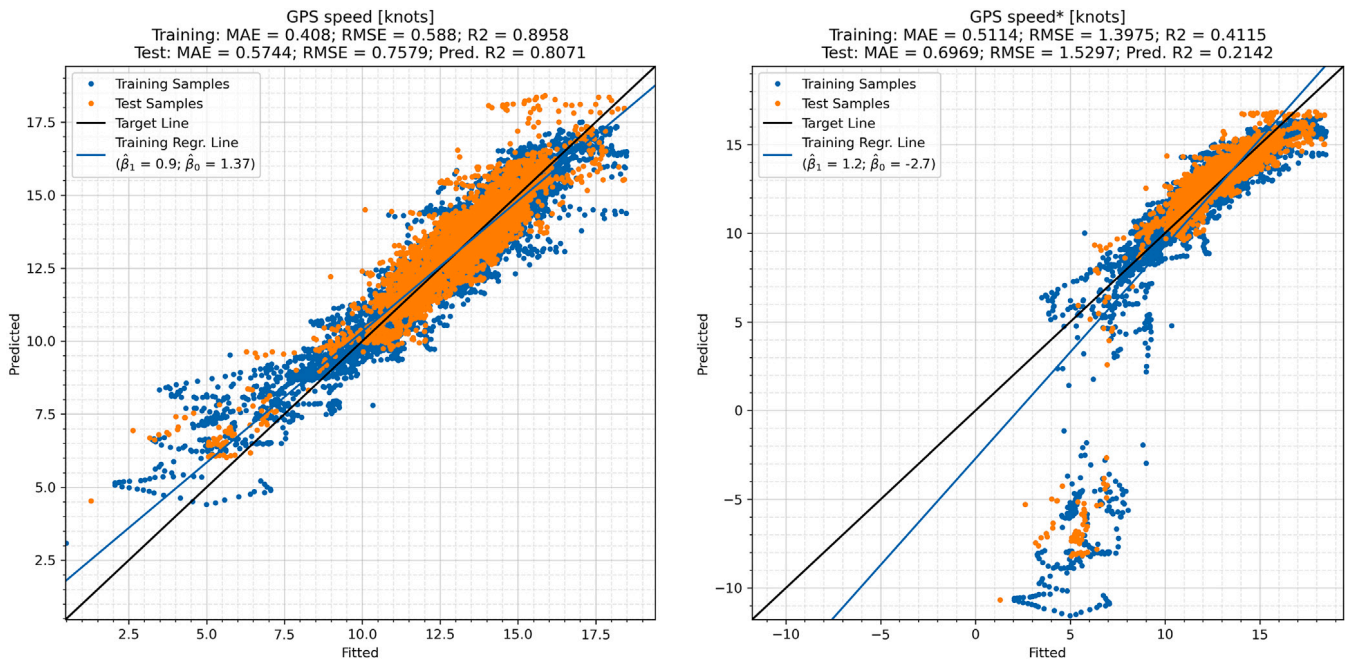

**Fig. 8.** Calibration result plots for linear and non-linear GPS speed variables using NL-PCR model for the sister ship. (For interpretation of the references to color in this figure legend, the reader is referred to the web version of this article.)

trends in the speed and power variables, an input data time-series, representing the assumed scenario, is fabricated and predictions are obtained from all the calibrated models.

Fig. 9 presents the predicted performance trends for both the ships from NL-PCR and NL-PLSR models, and Fig. 10 presents the trends predicted by the ANN model. The red and dashed black vertical lines represent the propeller and hull cleaning events, respectively. In all the cases, the shaft power predictions shows the expected trend with a substantially big drop in power demand, i.e., a substantial hydrodynamic performance improvement, just after the hull cleaning event in the case of the sister ship. The GPS and log speeds also shows the expected trends but only in the case of the sister ship. For the original ship, the NL-PCR and ANN models did not predict the expected trend for the ship speed variables and rather shows the opposite trend indicating a

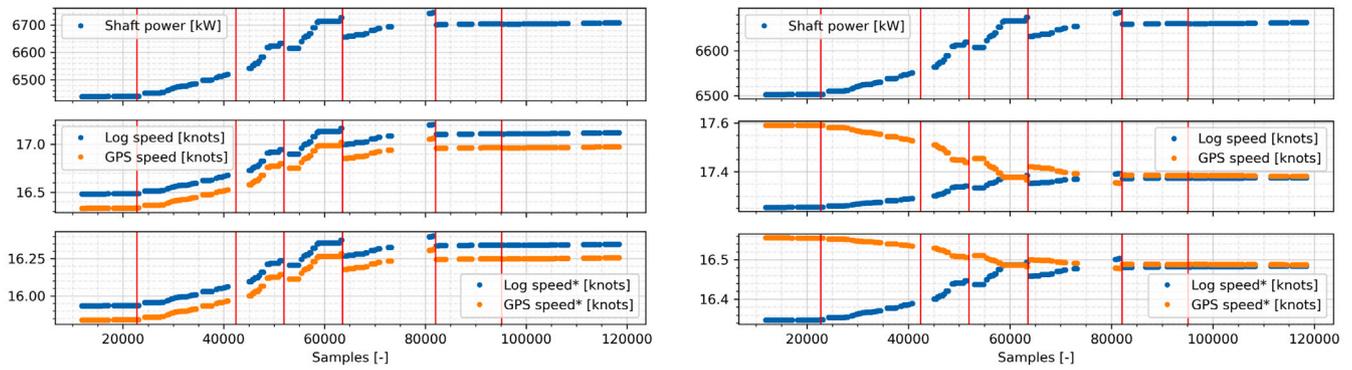

(a) Performance trends predicted by NL-PCR (left) and NL-PLSR (right) model for the original ship.

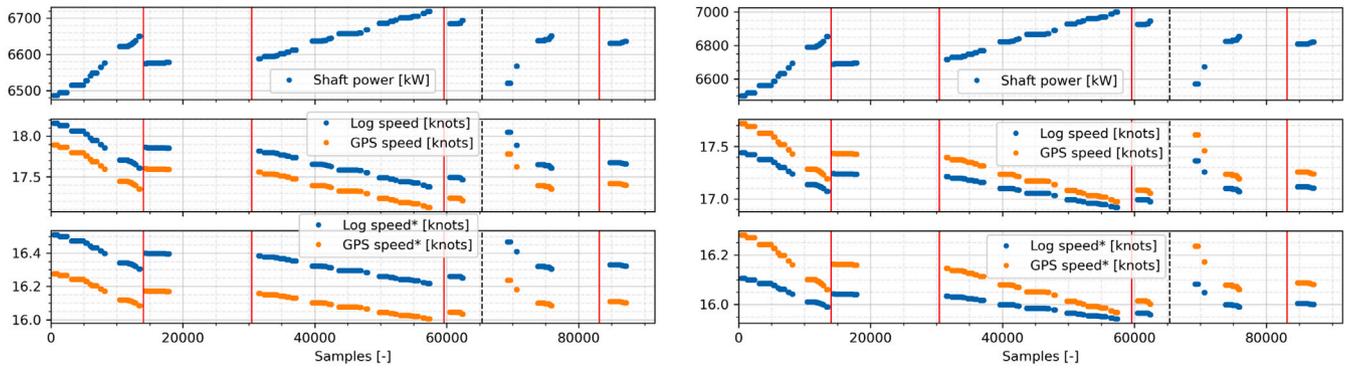

(b) Performance trends predicted by NL-PCR (left) and NL-PLSR (right) model for the sister ship.

**Fig. 9.** Performance trends predicted by NL-PCR (left) and NL-PLSR (right) model for the original ship. (For interpretation of the references to color in this figure legend, the reader is referred to the web version of this article.)

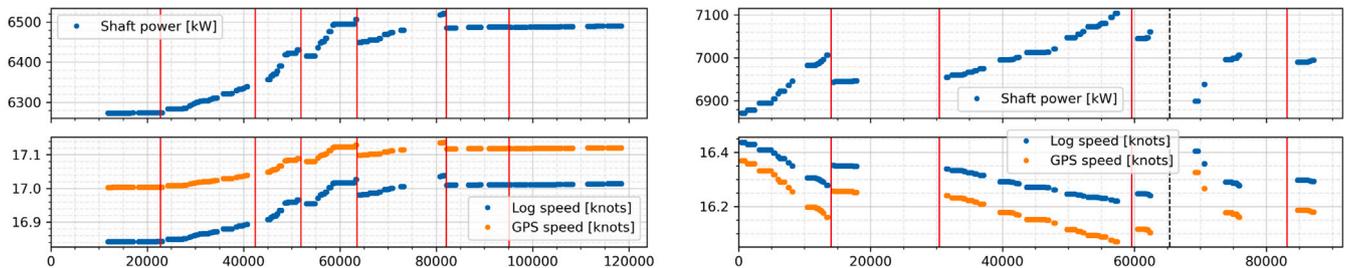

**Fig. 10.** Performance trends predicted by ANN model for the original ship (left) and the sister ship (right). The red and dashed black vertical lines represents the propeller and hull cleaning events, respectively. (For interpretation of the references to color in this figure legend, the reader is referred to the web version of this article.)

drop in hydrodynamic performance after each cleaning event, but the NL-PLSR model shows the expected trend using the GPS speed variable.

### 5.4. Ship performance monitoring & calm-water curves

The performance of a sea going ship is generally evaluated with respect to a reference point on its calm-water speed-power curve. For example, the drop in hydrodynamic performance of a ship has been reported either in terms of an increase in power demand for a given ship speed (Walker and Atkins, 2007; Ejdfors, 2019) or sometimes in terms of the speed-loss at a given shaft power (Koboević et al., 2019; Coraddu et al., 2019b). Similar estimates can be predicted using the calibrated machine-learning (ML) models to quantify the hydrodynamic performance of the ships. The current section presents the results in terms of the change in power demand of the ships, at the service speed, predicted through each propeller and hull cleaning event as well as for the starting and end of the time-series. The latter represents the change in hydrodynamic performance of the ships though the entire data recording duration. Here, the predicted change in power demand is calculated as the difference between the shaft power predicted for just before and after the corresponding event at the service speed of the ship.

Table 6 shows the predicted change in the power demand at the service speed for all the cleaning events and for the starting and end of the time-series for both the ships. A reduction in power demand (negative value with green cell color) indicates an improvement in the hydrodynamic performance of the ship whereas an increase in power demand (positive value with reddish cell color) indicates performance degradation. As a reference, the change in power demand for each case is also estimated using the fouling friction coefficient ($\Delta C_F$), presented in the rightmost column in Table 6.

The predicted change in power demand (presented in Table 6) for the sister ship shows an improvement in the hydrodynamic performance in all the cases except for the 2nd propeller cleaning event (*Prop. 2*). This may be due to a large section of unavailable data in the time-series just before the 2nd propeller cleaning event (as shown in Fig. 9).

**Table 6**

Performance prediction results showing the predicted change in power demand (in kW), through all the cleaning events and the starting and end of the time-series, using the regression models for both the ships. 'Prop.' stands for Propeller. The * marked target variables are the non-linear transformed variables (variable numbers 17 and 18 in Table 3) used in NL-PCR and NL-PLSR models. The green colored cells indicate performance improvement and the reddish ones indicate drop in performance.

| Cleaning Event | Change in Power demand [kW] | | | | | | | | | | |
| --- | --- | --- | --- | --- | --- | --- | --- | --- | --- | --- | --- |
| | NL-PCR | | | | NL-PLSR | | | | ANN | | Using $\Delta C_F$ |
| | Using GPS speed | Using Log speed | Using GPS speed* | Using Log speed* | Using GPS speed | Using Log speed | Using GPS speed* | Using Log speed* | Using GPS speed | Using Log speed | |
| **Original Ship** | | | | | | | | | | | |
| Prop. 1 | 0.01 | 0.01 | 0.01 | 0.01 | −0.01 | 0.00 | −0.01 | 0.00 | 0.00 | 0.00 | −716.03 |
| Prop. 2 | −27.31 | −24.92 | −27.52 | −25.46 | 28.22 | 4.81 | 21.42 | −1.97 | −0.20 | −3.73 | −247.98 |
| Prop. 3 | 23.58 | 21.52 | 24.59 | 20.21 | −24.37 | −4.16 | −18.47 | 1.79 | −5.23 | −1.55 | −930.51 |
| Prop. 4 | 86.83 | 79.23 | 83.43 | 76.64 | −89.74 | −15.31 | −67.95 | 6.85 | −20.31 | −7.29 | −652.87 |
| Prop. 5 | 54.21 | 49.46 | 52.77 | 48.49 | −56.03 | −9.56 | −42.40 | 4.35 | −12.99 | −4.89 | −274.52 |
| Prop. 6 | 1.34 | 1.22 | 1.29 | 1.19 | −1.38 | −0.24 | −1.05 | 0.11 | −0.32 | −0.12 | 534.02 |
| After 3 yrs | −330.58 | −301.68 | −329.82 | −299.47 | 341.72 | 58.25 | 259.16 | −24.45 | 33.74 | −14.62 | −464.61 |
| **Sister Ship** | | | | | | | | | | | |
| Prop. 1 | −308.67 | −308.25 | −220.64 | −232.30 | −390.39 | −324.00 | −332.12 | −253.17 | −192.33 | −158.05 | 777.26 |
| Prop. 2 | 39.00 | 39.00 | 27.83 | 29.49 | 49.39 | 40.99 | 41.52 | 32.06 | 24.79 | 20.49 | 291.27 |
| Prop. 3 | −137.30 | −137.25 | −101.66 | −102.05 | −173.44 | −143.95 | −150.14 | −112.16 | −98.60 | −74.08 | 32.90 |
| Hull & Prop. 4 | −703.66 | −702.93 | −508.24 | −529.98 | −889.62 | −738.71 | −759.10 | −577.31 | −454.95 | −374.73 | −1286.20 |
| Prop. 5 | −82.89 | −82.68 | −60.06 | −62.09 | −104.71 | −86.91 | −91.14 | −67.85 | −53.17 | −43.21 | 313.39 |
| After ~2.5 yrs | 605.29 | 605.01 | 432.85 | 458.78 | 765.45 | 635.93 | 647.93 | 497.69 | 408.91 | 348.46 | 568.17 |

Moreover, as expected, the biggest improvement is predicted for the hull and propeller cleaning event (*Hull & Prop. 4*). For the original ship, the values predicted by the NL-PCR model shows performance degradation in all the cases except for the 2nd propeller cleaning event (*Prop. 2*). This is probably due to the fact that the NL-PCR model mapped a performance degradation trend for speed variables, as shown in Fig. 9. The ANN model also predicted a performance degradation trend for speed variables, but it still managed to predict a very small performance improvement for almost all the cleaning events. Here, the NL-PLSR model predicts the highest performance improvement using the linear and non-linear (* marked) GPS speed variables, due to favorable trends shown for these two variables in Fig. 9.

Comparing with the change in power demand predicted using the fouling friction coefficient (*ΔC_F*), all the values predicted by the machine-learning (ML) models are highly under-estimated except for the starting and end of the time-series for the sister ship (*After ~2.5 yrs*), only in this case the values seems to be in the same range. But the values obtained using $\Delta C_F$ shows unexpected drop in the hydrodynamic performance of the ship for all the propeller cleaning events for the sister ship and the last propeller cleaning event (*Prop. 6*) for the original ship. Moreover, the change in power demand predicted for the starting and end of the time-series for the original ship (*After 3 yrs*) using $\Delta C_F$ shows improvement in the performance of the ship, which is highly unexpected. Finally, using $\Delta C_F$, a drop in power demand by about 1300 kW is predicted for the hull and propeller cleaning event for the sister ship (*Hull & Prop. 4*) whereas the values predicted using the ML models is in the range of 375 to 890 kW, of which the ANN model predicted the lowest and the NL-PLSR model predicted the highest.

Overall, a clear assessment of the quality of the results cannot be obtained here as the traditional method (using $\Delta C_F$) itself does not seem to be predicting the expected results in about half of the cases, and there is no alternative method available to make a better quantitative assessment. Nevertheless, the results obtained from all the ML models seems to be qualitatively good, at least in the case of the sister ship. This may be attributed to the fact that the in-service data obtained from the original ship, being newly-built around the data recording duration, has a small correlation with the fouling growth factor (FGF), but the sister ship has a stronger correlation with the FGF, and therefore, shows better results using the ML models.

Further, it is also possible to predict the whole calm-water speed-power curve for the ships for before and after each of the propeller and hull cleaning events as well as for the starting and end of the time-series. Predicting the calm-water speed-power curve would also act as a generalization check, showing the underlying mapping to which each machine-learning (ML) model is fitted. Moreover, the predicted curve can be compared with experimentally obtained calm-water speed-power curves, for example, the model test and sea-trial results. Fig. 11 shows the improvement in hydrodynamic performance of the sister ship by predicting the calm-water speed-power curves (for ballast draft and trim) before and after the hull and propeller cleaning event (*Hull & Prop. 4*) using all the three models. The values in Table 6 are actually the distance between these predicted calm-water curves along the shaft power axis at the service speed of the ship.

In Fig. 11, the speed-power curves, obtained using the NL-PCR and NL-PLSR models, are cubic in nature due to the non-linear (cubic) transformation adopted for the speed variables. It is expected that the post-cleaning curve should almost match with the sea-trial curve (in Fig. 11), but this is not observed here. This may be attributed to the fact that the sea trial curve (presented in Fig. 11) are obtained for another sister ship (not the here referred sister ship). Nevertheless, the shifting of the calm-water speed-power curve (pre- and post-cleaning) shows a substantial improvement in the hydrodynamic performance of the ship. Further, it is observed that the predicted curves do not pass through the origin (0, 0), most definitely due to the sparsity of data in the lower speed-power range. The probabilistic ANN model also predicts a higher uncertainty (presented by the shaded region around the predicted points in Fig. 11) for the lower speed-power range due to the same reason. A better curve may be obtained using a more balanced dataset.

Fig. 12 shows the calm-water speed-power curves predicted for the starting and end of the time-series for the original ship using the ANN model. Here, the curve is predicted using the GPS speed variable, but it should match with the log speed variable as the water current is assumed to be absent while making these predictions. In this case, it is observed that the speed-power curve is well below the sea-trial and model test curves. This may be due to an extrapolation on the draft axis as the in-service data contains a very small amount of data in the ballast daft range, in the case of the original ship, as evident from Fig. 13. Again, a better curve may be obtained using a more balanced dataset.

## 6. Conclusion

The current work presents a novel method to monitor the hydrodynamic performance of a sea-going ship using the in-service data recorded onboard the ship. The in-service data from two sister ships, complemented with weather hindcast data, is used to calibrate three machine-learning (ML) models, namely, non-linear Principal Component Regression (NL-PCR), non-linear Partial Least Squares Regression (NL-PLSR) and probabilistic artificial neural network (ANN), through several propeller and hull cleaning events. A fouling growth factor (FGF) is included in the ML models to incorporate the effect of fouling growth on the hull and propeller of the ships. The FGF is formulated using the ships' static time in water and the fouling trends obtained using the generalized admiralty coefficient for the ships.

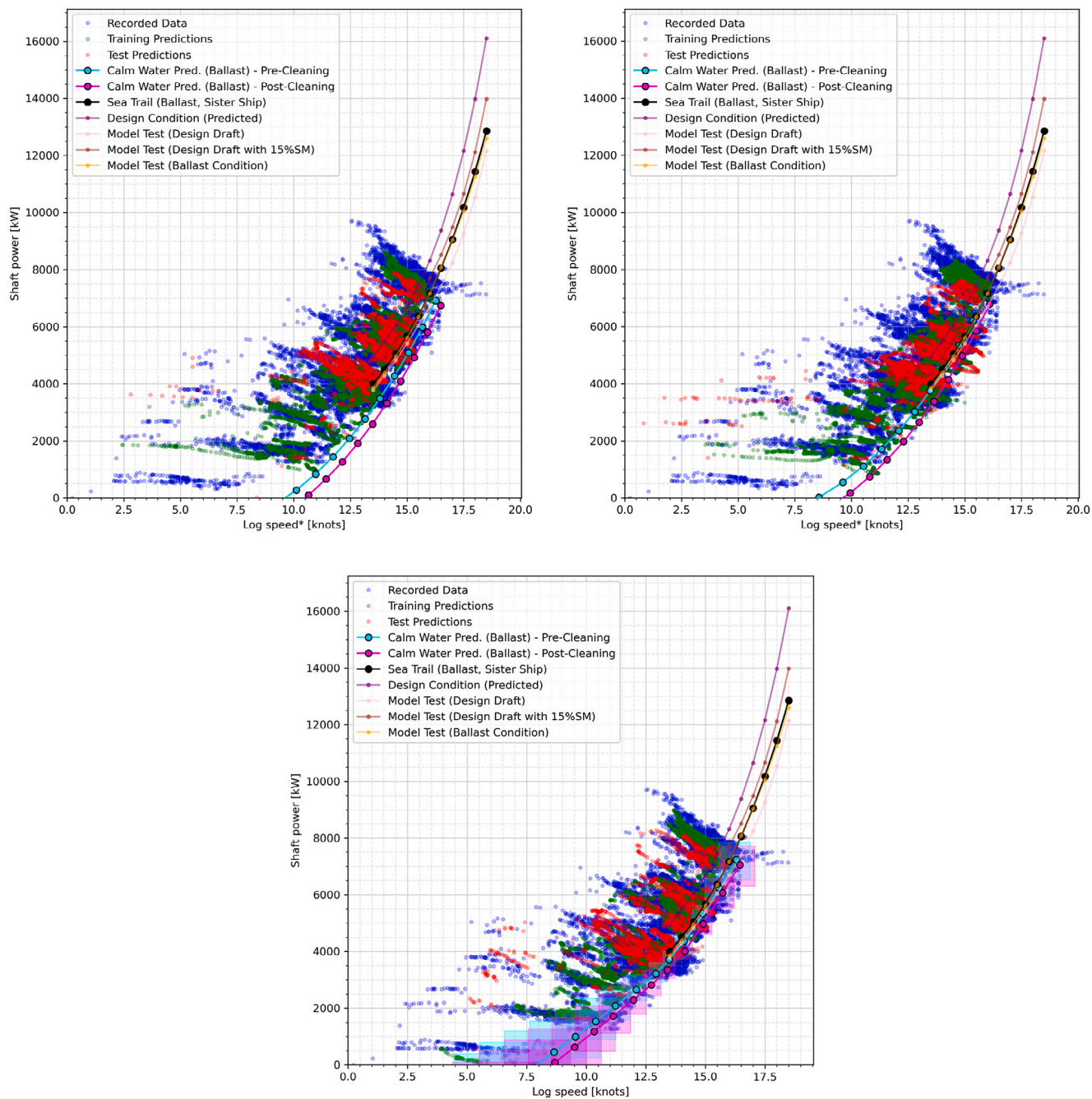

**Fig. 11.** Calm-water speed-power curves predicted by NL-PCR (left), NL-PLSR (center) and probabilistic ANN (right) models for the sister ship for just before (pre-cleaning) and after (post-cleaning) the hull and propeller cleaning event (*Hull & Prop. 4*). The shaded region around the predicted calm-water curve, in case of the ANN model, presents the 95% confidence interval estimated using the approximate predictive distributions predicted for each sample. (For interpretation of the references to color in this figure legend, the reader is referred to the web version of this article.)

The results indicate that it may be possible to use simple interpretable ML models, like NL-PCR and NL-PLSR, instead of a highly complex black-box model, like ANN, to model the hydrodynamic state of a sea-going ship. PCR and PLSR are basically linear models but empowered by some approximate non-linear transformations, obtained from our domain knowledge, produced results comparable to the ANN model in the given case. These models can be further improved by using a set of more accurate non-linear transformations. Further, it is also shown in the current work that it may be necessary to use a balanced dataset to predict an accurate calm-water speed-power curve over the whole range of speed and power. The in-service datasets used here are sparse in the lower speed range, as is generally seen in most of the ship propulsion datasets, resulting in a poor prediction in that range.

The fouling friction coefficient ($\Delta C_F$) fails as a reference for validation as the change in performance predicted by using $\Delta C_F$ does not seem to be valid in about half of the cases. Thus, the change in performance predicted by the ML models could not be validated here, but the results for at least one ship indicate an improvement in the hydrodynamic performance of the ship for almost all the propeller and hull cleaning events, with the highest improvement predicted for the hull cleaning event, which is as expected.

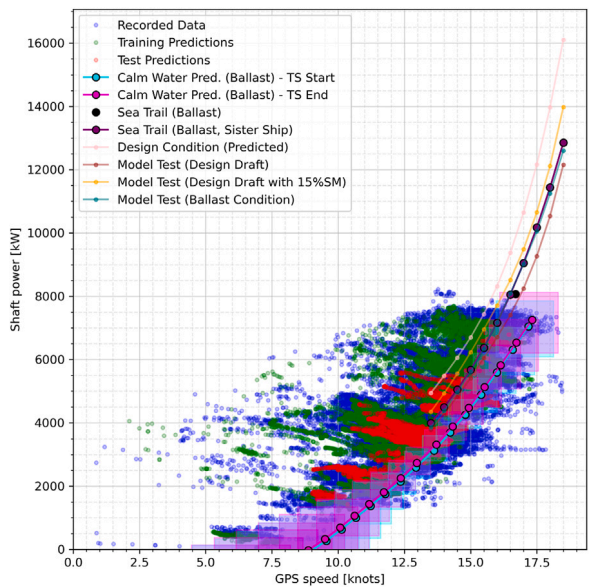

**Fig. 12.** Calm-water speed-power curve predicted (using the GPS speed variable) by probabilistic ANN model for the original ship for the starting and end of the time-series (*After 3 yrs*). The shaded region around the predicted calm-water curve presents the 95% confidence interval estimated using the approximate predictive distributions predicted for each sample. (For interpretation of the references to color in this figure legend, the reader is referred to the web version of this article.)

## 7. Future work

As compared to the fouling growth model presented by Malone et al. (1981), the fouling growth factor (FGF) formulated here is quite simplified. Most importantly, it is assumed here that the fouling grows at the same rate for each unit static time between any two cleaning events, reflected by using the same fouling growth rate (FGR) in Eq. (11), for all the samples between two cleaning events. The FGR here is estimated as the slope of the mean generalized admiralty coefficient (shown in Fig. 3). In a more accurate approach, the FGR should be estimated for each individual port visit as the FGR depends on the water conditions in which the ship is static, explained in Section 3. Due to substantial noise in the mean generalized admiralty coefficient values (as seen in Fig. 3), it is not possible to estimate a set of logically valid FGRs for each port visit using this approach. Thus, an alternate method or approach needs to be defined to get a better FGF.

On another note, it may also be a good idea to use a different set of machine-learning (ML) models for the given case. Although ANN is still considered as one of the best ML models, there are several other ML models which are known to have outperformed ANN in different scenarios. Moreover, the probabilistic ANN model used here is an approximation for a true Bayesian model, a full Bayesian approach would probably produce better estimation of the uncertainties in the predictions. It would, therefore, be interesting to see if a different ML model can be used to solve this problem more efficiently. Finally, the current work is only validated for two sister ships. It would most definitely be desired to validate this approach on a more variant group of datasets.

## CRediT authorship contribution statement

**Prateek Gupta:** Conceptualization, Methodology, Investigation, Software, Data curation, Visualization. **Adil Rasheed:** Conceptualization, Supervision, Writing. **Sverre Steen:** Conceptualization, Supervision, Resources.

## Declaration of competing interest

The authors declare that they have no known competing financial interests or personal relationships that could have appeared to influence the work reported in this paper.

## Acknowledgments

This study is part of the research project SFI Smart Maritime – Norwegian Centre for Improved Energy-Efficiency and Reduced Emissions from the Maritime Sector (RCN project number 237917).

## Appendix A. Validation of generalized admiralty coefficient as a ship's hydrodynamic performance indicator

Gupta et al. (2021) proposed to use a generalized form of admiralty coefficient as a statistical performance indicator for a ship. The paper presented the validation of the method for a new-built ship (here referred to as the original ship). This appendix present the validation of the same method using an extended dataset from the original ship as well as for its sister ship. It should be noted that the data obtained from the sister ship is recorded after a few years of service and includes a hull cleaning (dry-docking) event whereas the data from the original ship only includes propeller cleaning events. Thus, the sister ship is expected to show a greater extent of fouling.

First of all, a generalized admiralty coefficient is obtained, in much the same way as presented by Gupta et al. (2021), for each ship by statistically fitting the generalized admiralty coefficient formula ($\Delta^m V^n / P_s$) to the near-calm-water filtered and corrected in-service data, and then, the fouling growth rate is estimated as the trend in the obtained generalized admiralty coefficient for all the legs between each propeller and/or hull cleaning events. Finally, in order to validate the method, the fouling growth rate predicted by the obtained generalized admiralty coefficient is compared, qualitatively, with the fouling growth rate predicted by the traditional method, i.e., the fouling friction coefficient ($\Delta C_F$).

The near-calm-water in-service data is obtained by, first, filtering the recorded in-service data for small wind and wave load conditions (total wind speed <5.5 m/s and significant wave height <1 m) and, then, correcting the shaft power measurements in the filtered dataset for wind and wave loads. The wind loads are estimated using Fujiwara's method (Fujiwara et al., 2005), as recommended by ITTC (2017), and the wave loads are estimated using Liu and Papanikolaou's method (Liu and Papanikolaou, 2020). Unlike DTU's method (Martinsen (2016), Taskar and Andersen (2021); used by Gupta et al. (2021)), which provides added wave resistance estimates for mean wave heading from head seas to beam seas only, Liu and Papanikolaou's method (Liu and Papanikolaou, 2020), used here, provides the estimated added wave resistance for all the wave headings, including the following waves.

Figs. 14 and 15 presents the validation of the generalized admiralty coefficient method for both the original ship (also presented by Gupta et al. (2021)) as well as its sister ship.

## Appendix B. Calculating fouling friction coefficient ($\Delta C_F$)

The fouling friction coefficient ($\Delta C_F$) is calculated as the difference between the total resistance coefficient ($C_{T,Data}$) obtained from the in-service data in calm-water conditions and the total resistance coefficient ($C_{T,Emp}$) obtained from well-established empirical methods or model test results.

$$\Delta C_F = C_{T,Data} - C_{T,Emp} \qquad (12)$$

The total resistance coefficient ($C_{T,Data}$) from in-service data can be obtained as follows:

$$C_{T,Data} = \frac{R_{T,Data}}{0.5\rho S V^2} = \frac{P_s}{0.5\rho S V^3 \eta_T} \qquad (13)$$

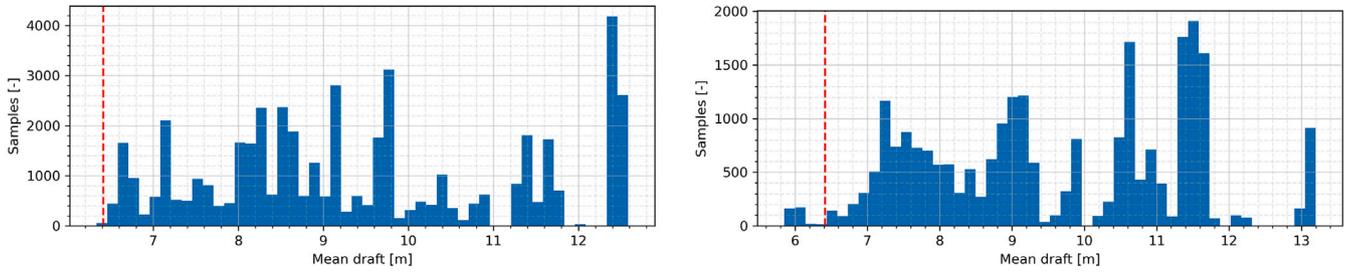

**Fig. 13.** Distribution of mean draft for the original ship (left) and the sister ship (right). The dashed red vertical line indicates the ballast draft of the ship.

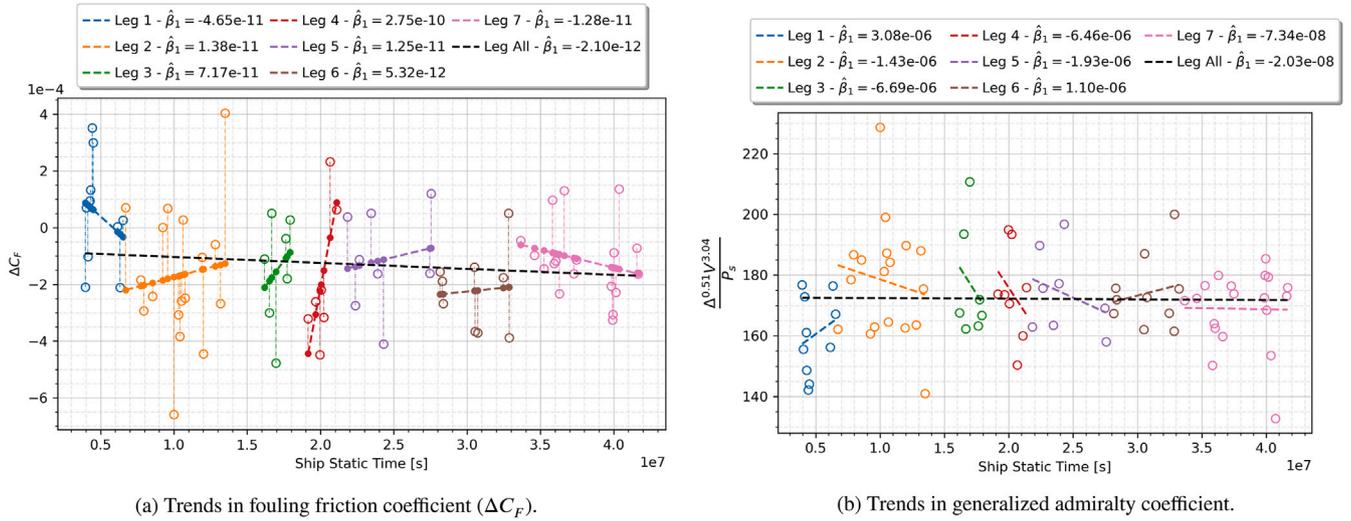

(a) Trends in fouling friction coefficient ($\Delta C_F$).

(b) Trends in generalized admiralty coefficient.

**Fig. 14.** Trends in fouling friction coefficient ($\Delta C_F$) and generalized admiralty coefficent for original ship. (For interpretation of the references to color in this figure legend, the reader is referred to the web version of this article.)

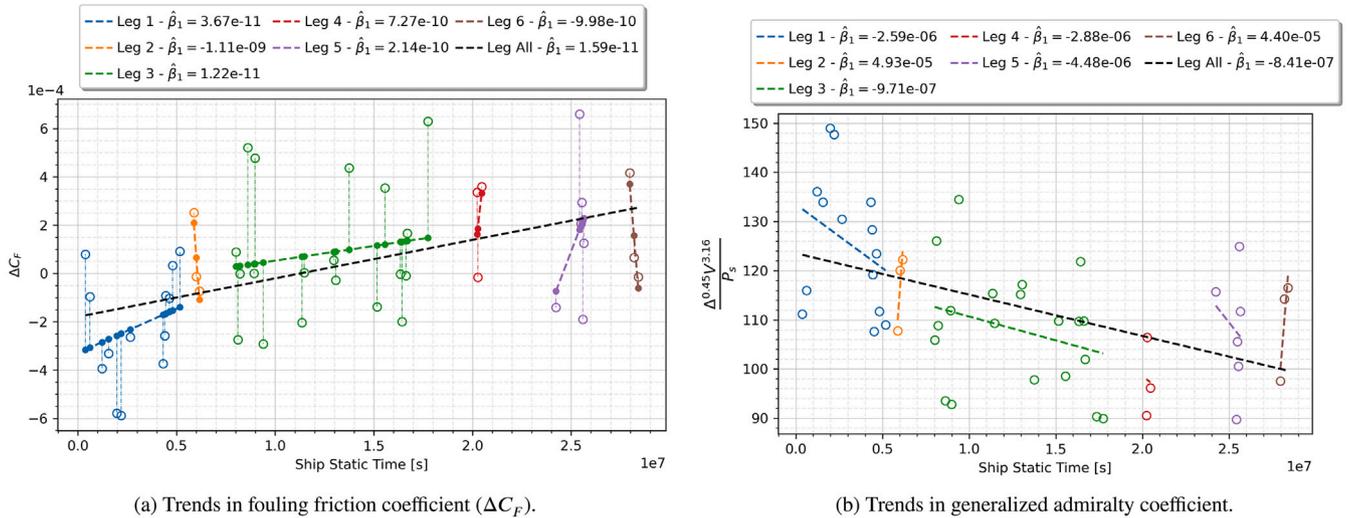

(a) Trends in fouling friction coefficient ($\Delta C_F$).

(b) Trends in generalized admiralty coefficient.

**Fig. 15.** Trends in fouling friction coefficient ($\Delta C_F$) and generalized admiralty coefficent for sister ship. (For interpretation of the references to color in this figure legend, the reader is referred to the web version of this article.)

Where $R_{T,Data}$ is the total resistance (from data), $\rho$ is the density of sea water, $S$ is the wetted surface area (WSA), $V$ is the log speed (measured), $P_s$ is the shaft power (measured) and $\eta_T$ is the total propulsive efficiency. Here, the WSA can be estimated using the hull form (or offset table) of the ship for the corresponding mean draft and trim. The total propulsive efficiency ($\eta_T$) can be estimated empirically or using the model test results. In the above formula, special attention should be paid towards the units of contributing variables, for example, the unit of log speed should be m/s if the unit of shaft power is watt or kilowatt.

Further, the empirically obtained total resistance coefficient ($C_{T,Emp}$) can be dividing into individual resistance components as follows:

$$C_{T,Emp} = \frac{R_{T,Emp}}{0.5\rho SV^2} = \frac{R_{Calm} + R_{Wind} + R_{Wave} + R_{Others}}{0.5\rho SV^2} \quad (14)$$

Where $R_{T,Emp}$ is the total resistance (from empirical methods), $R_{Calm}$ is the calm water resistance, $R_{Wind}$ is the added wind resistance, $R_{Wave}$ is the added wave resistance and $R_{Others}$ is the resistance from other effects. These individual resistance components can be further estimated using different physics-based and/or empirical methods. $R_{Others}$ can include the effect of transom stern, appendages and so on.